\documentclass[letterpaper, 10 pt, journal, twoside]{IEEEtran}  
\IEEEoverridecommandlockouts         

\usepackage{graphics}
\usepackage[pdftex]{graphicx}
\usepackage[tight,footnotesize]{subfigure}
\usepackage{multirow}
\usepackage[cmex10]{amsmath}
\usepackage{amssymb}
\usepackage{amsfonts}
\usepackage{algpseudocode}
\usepackage{algorithm}
\usepackage{booktabs}
\usepackage{textcomp}
\usepackage{url}
\usepackage{color}
\usepackage{siunitx}
\usepackage{soul}

\usepackage{wrapfig}
\usepackage{babel}
\usepackage[small]{caption}

\usepackage{multirow}
\usepackage{wrapfig}
\usepackage{colortbl}
\usepackage[normalem]{ulem}

\newcolumntype{g}{>{\columncolor{CuGray}}c}
\newcolumntype{z}{>{\columncolor{CuGray}}l}

\renewcommand{\paragraph}[1]{\noindent\textbf{#1.}\,\,}

\usepackage{xspace}

\def\onedot{.\@\xspace}
\def\eg{\emph{e.g}\onedot} 
\def\ie{\emph{i.e}\onedot}

\def\wrt{\emph{w.r.t}\onedot} 
\def\etal{\emph{et al}\onedot}

\newcommand{\Sref}[1]{Sec.~\ref{#1}}

\newcommand{\Fref}[1]{Fig.~\ref{#1}}
\newcommand{\Tref}[1]{Table~\ref{#1}}

\newcommand{\AiSDF}{{\texttt{AiSDF}}}

\newcommand{\be}{\begin{eqnarray}}
\newcommand{\ee}{\end{eqnarray}}
\newcommand{\bee}{\begin{eqnarray*}}
\newcommand{\eee}{\end{eqnarray*}}

\newcommand{\matrixb}{\left[ \begin{array}}
\newcommand{\matrixe}{\end{array} \right]}

\newcommand{\argmin}{\operatornamewithlimits{\arg \min}}

\title{
AiSDF: Structure-aware Neural Signed Distance Fields in Indoor Scenes
}

\author{Jaehoon Jang$^{1*}$, Inha Lee$^{1*}$,  Minje Kim$^{1}$ and Kyungdon Joo$^{2\dag}$ \\
\thanks{Manuscript received: September 19, 2023; Revised January 17, 2024; Accepted February 20, 2024.
This paper was recommended for publication by Associate Editor T. Asfour and Editor C. Cadena Lerma upon evaluation of the Associate Editor and Reviewers' comments.
This work was supported by Institute of Information \& communications Technology Planning \& Evaluation (IITP) grant funded by the Korea government (MSIT) (No.2022-0-00907, Development of AI Bots Collaboration Platform and Self-organizing, No.2020-0-01336, Artificial Intelligence Graduate School Program (UNIST)).}
\thanks{$^{1}$J. Jang, I. Lee and M. Kim are with the Artificial Intelligence Graduate School, UNIST, Ulsan, South Korea. {(\tt\footnotesize e-mail: erick1997@unist.ac.kr; epsilon8854@unist.ac.kr; minje617@unist.ac.kr})}
\thanks{$^{2}$Kyungdon Joo is with the Artificial Intelligence Graduate School and the Department of Computer Science and Engineering, UNIST, Ulsan, South Korea. {(\tt\footnotesize e-mail: kyungdon@unist.ac.kr})}%
\thanks{$^{*}$Equal contribution (alphabetical order by last name)}
\thanks{$^{\dag}$Corresponding author}
\thanks{Project website: https://vision3d-lab.github.io/AiSDF/}
\thanks{Digital Object Identifier (DOI): see top of this page.}
}

\begin{document}

\maketitle

\begin{figure*}[!]
    \vspace{-3mm}
    \centering
    \captionsetup{type=figure}
    \includegraphics[width=0.85\linewidth]{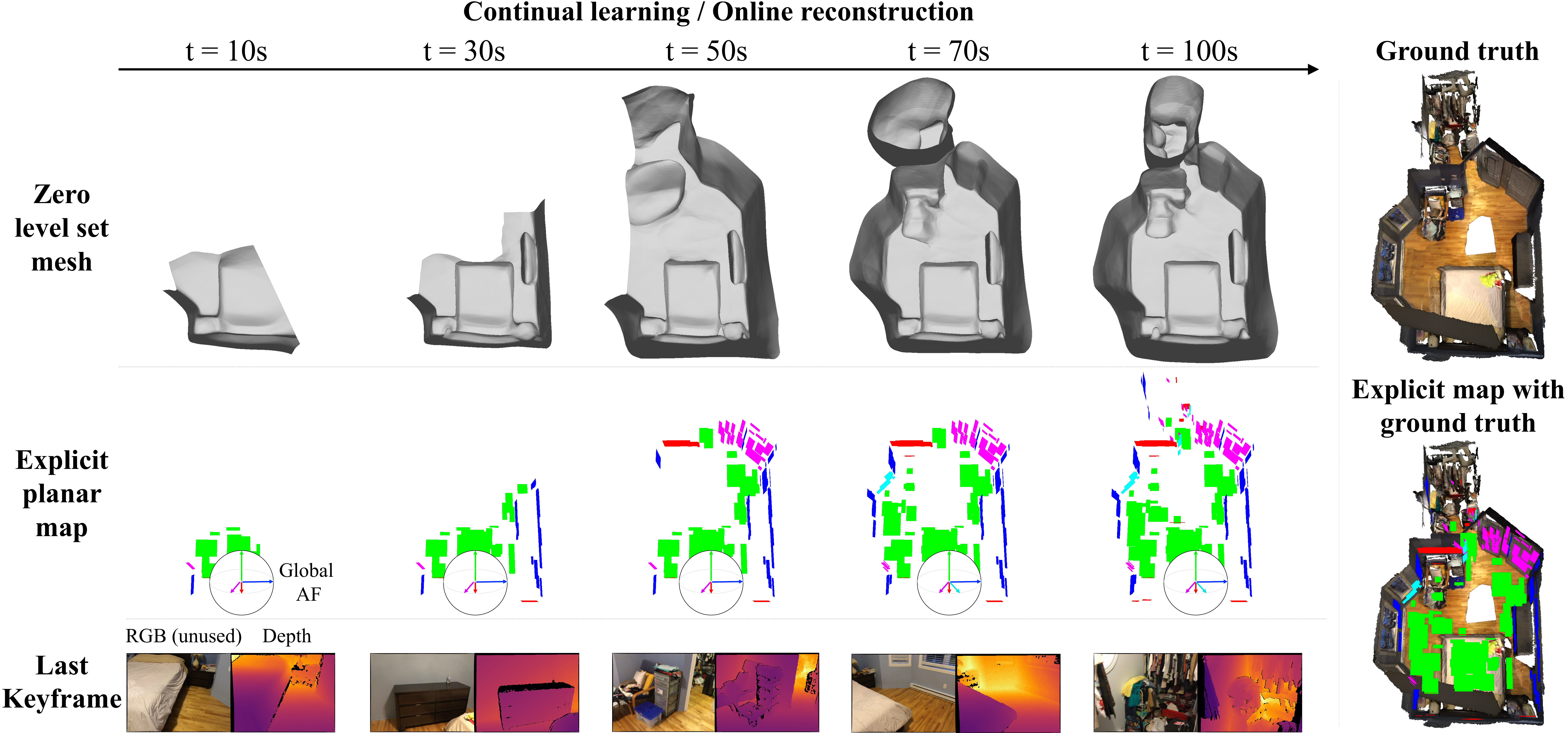}
    \captionof{figure}{
     \textbf{The proposed {\AiSDF} on the ScanNet dataset~\cite{dai2017scannet}}. 
    \emph{Top}: Our framework represents the scene as a signed distance fields (SDF) by considering the structure of the scene in a continual manner. 
    \emph{Middle}: In addition, we estimate the underlying Atlanta structure (global Atlanta frame) and extract a 3D explicit planar map in the form of Atlanta-aware surfels.
    We colorize each surfel with the associated Atlanta direction.
    \emph{Bottom}: We visualize the last keyframe used by {\AiSDF} (RGB image is unused in practice).    }
    \label{fig:teaser}
    \vspace{-3mm}
\end{figure*}

\begin{abstract}
    Indoor scenes we are living in are visually homogenous or textureless, while
    they inherently have structural forms and provide enough structural priors for
    3D scene reconstruction. 
    Motivated by this fact, we propose a structure-aware online signed distance fields (SDF) reconstruction framework in indoor scenes, especially under the Atlanta world (AW) assumption. 
    Thus, we dub this incremental SDF reconstruction for AW as AiSDF. 
    Within the online framework, we infer the underlying Atlanta structure of a given scene and then estimate planar surfel regions supporting the Atlanta structure. 
    This Atlanta-aware surfel representation provides an explicit planar map for a given scene. 
    In addition, based on these Atlanta planar surfel regions, we adaptively sample and constrain the structural regularity in the SDF reconstruction, which enables us to improve the reconstruction quality by maintaining a high-level structure while enhancing the details of a given scene. 
    We evaluate the proposed AiSDF on the ScanNet and ReplicaCAD datasets, where we
    demonstrate that the proposed framework is capable of reconstructing fine details
    of objects implicitly, as well as structures explicitly in room-scale scenes.
\end{abstract}


\section{Introduction}  \label{sec:intro}
\IEEEPARstart{V}{arious} 3D scene representations, such as explicit geometric primitive and implicit functions, have been actively studied in computer vision and robotics~\cite{cabral2014piecewise,schops2019surfelmeshing,Mittal_2022_CVPR}.
As one of the implicit representations, signed distance fields (SDF) inherently encode the surface information as the signed distance between the position in space and the closest surface, where the zero-level set corresponds to the surface. 
By virtue of this characteristic, many vision tasks, such as rendering~\cite{Jiang_2020_CVPR}, and path planning~\cite{zucker2013chomp}, use SDF as a medium, especially neural implicit reconstruction based on SDF has gained a lot of attention~\cite{sucar2021imap, zhu2022nice}. 
In those tasks, inferring accurate SDF with low latency is important.

Recently, Ortiz~\etal~\cite{iSDF2022} presented an incremental SDF estimation framework (iSDF in short) that reconstructs the SDF of room-scale indoor environments using continual learning in real-time.
Given a stream of posed depth images, iSDF focuses on a neural SDF-based mapping module within a SLAM framework.
By employing a compact MLP network and sparse sampling, they show that online SDF reconstruction is feasible using continual learning.
However, iSDF misses several properties in \emph{indoor scenes} that can improve reconstruction quality while maintaining efficiency.

From the layout or floorplan of the rooms to various objects such as furniture, most objects, including the scene itself,  have structural forms in indoor scenes. 
Concretely, while they are homogeneous or textureless from a visual perspective, they consist of a set of orthogonal or parallel planes (planar segments) from a geometric viewpoint. 
These structural characteristics of indoor scenes can be represented by a few dominant directions; structural assumptions, such as the Manhattan world (MW)~\cite{coughlan1999manhattan}, or Atlanta world (AW)~\cite{SchindlerDellaert04} assumptions, have been explored in the literature~\cite{gupta2013perceptual,carlone2015initialization,kim2018low,joo2021linear}. 
For example, the MW assumption, represented by three orthogonal directions, describes a given scene with a strictly aligned shape, like a cuboid shape. In the case of the AW assumption, it can cover more general indoor scenes, such as non-orthogonal walls, using vertical and a set of horizontal directions. 
These structural assumptions have been exploited as prior information on indoor scenes.

Motivated by this fact, we propose a structure-aware online SDF reconstruction framework, \AiSDF, in indoor scenes under the AW assumption (see \Fref{fig:teaser}).
To this end, we continually estimate the underlying Atlanta structure of a given scene inside the online SDF reconstruction framework.
This structural understanding provides several advantages within our {\AiSDF} framework.
1) Based on the estimated dominant Atlanta directions as a priori, we can efficiently extract planar regions following the AW assumption in the form of surface elements (surfels). 
This Atlanta-aware surfel representation provides an explicit planar map for a given scene.
2) We can exploit the structural regularity as a constraint. Concretely, we can adaptively sample points according to the Atlanta-aware surfels, which enables us to enforce the additional structural constraint and focus on complex regions.
We seamlessly integrate the structural understanding into the online SDF reconstruction framework.
We demonstrate our {\AiSDF} framework on the ScanNet~\cite{dai2017scannet} and ReplicaCAD~\cite{szot2021habitat} datasets. 
{\AiSDF} shows that the overall details of the scene and the structure in the form of a 3D planar map are recovered better than comparison methods~\cite{oleynikova2017voxblox, iSDF2022}.
In summary, our contributions are: 
\begin{itemize}
    \item We propose a new structure-aware online neural SDF, \texttt{AiSDF} that reconstructs a given indoor scene under the AW assumption with an online process.
    \item Based on the structural understanding, we introduce an Atlanta-aware surfel representation, which approximates a given indoor scene to a set of rectangular surfels.
    \item By utilizing the Atlanta-aware surfels, we effectively sample points considering the structure of the scene. 
    In addition, we perform a structure-aware tight bound computation for self-supervised learning.
    \item In addition to obtaining a neural implicit map, \texttt{AiSDF} extracts a low-memory explicit planar map that can make it easier for robots to access the structure information of the scene.
\end{itemize}

\section{Related Work}  \label{sec:related_work}

\noindent \textbf{Structural assumptions}. \ 
Thanks to their simplicity, represented by a few dominant directions, and their applicability in structured environments, various structural assumptions have been studied in robotics and computer vision~\cite{straub2017manhattan,joo2019globally,joo2021linear}~(please refer to \cite{straub2017manhattan} for a detailed review).
The Manhattan world (MW) assumption~\cite{coughlan1999manhattan}, represented by three orthogonal directions, can approximate cuboid shape scenes, such as an indoor room. 
Beyond the MW assumption, the Atlanta world (AW) assumption~\cite{SchindlerDellaert04} is defined by a vertical direction and a set of horizontal directions orthogonal to the vertical one, which can describe most indoor scenes, including non-orthogonal walls.

These two structural assumptions have been broadly used in various vision applications, such as scene understanding~\cite{silberman2012indoor,choi2013understanding}, camera calibration~\cite{wildenauer2012robust}, visual odometry~\cite{kim2018low}, SLAM~\cite{carlone2015initialization,joo2021linear}, and so on. 
In particular, Joo~\etal~\cite{joo2021linear} propose a linear SLAM framework for structured environments. 
They estimate the underlying Atlanta structure and explicitly use planar features supporting the Atlanta structure as measurements.
In this work, we assume a given indoor scene follows the AW assumption. 

\vspace{1mm} \noindent \textbf{Neural scene reconstruction}. \
Recently, research on neural scene reconstruction using implicit representations, such as occupancy, neural radiance fields (NeRF), and SDF has been actively conducted~\cite{mescheder2019occupancy, mildenhall2020nerf, park2019deepsdf, po2023instant}.
Among various implicit representations, SDF has gained much attention in that it can implicitly encode surface information using continuous values~\cite{murez2020atlas,sun2021neuralrecon, yan2021continual, dai2022neural}. 
In addition, SDF can make significant synergy with volumetric rendering and produce a high-quality reconstruction~\cite{yariv2021volume,wang2021neus, Azinovic_2022_CVPR}.

Neural scene reconstruction also can be utilized in conjunction with online processes, such as SLAM~\cite{sucar2021imap, zhu2022nice}.
Within the traditional real-time SLAM pipeline (\ie, front-end tracking and back-end mapping), iMAP~\cite{sucar2021imap} takes an RGB-D image as input and exploits an MLP to implicitly represent a 3D volumetric map in the form of volume density. 
Recently, Ortiz~\etal~\cite{iSDF2022} propose iSDF, a continual SDF reconstruction framework given a stream of posed depth images. 
Inspired by iMAP, iSDF adopts a keyframe selection module to achieve real-time performance as well as train the model in a continual manner.
Although this recent progress in online implicit 3D scene reconstruction is impressive and impactful, they less attention to improving the reconstruction quality. 
In other words, they focus on the efficiency of 3D neural scene representation.

\begin{figure*}[t]
    \vspace{-3mm}
    \centering
    \captionsetup{type=figure}
    \includegraphics[width=0.9\linewidth]{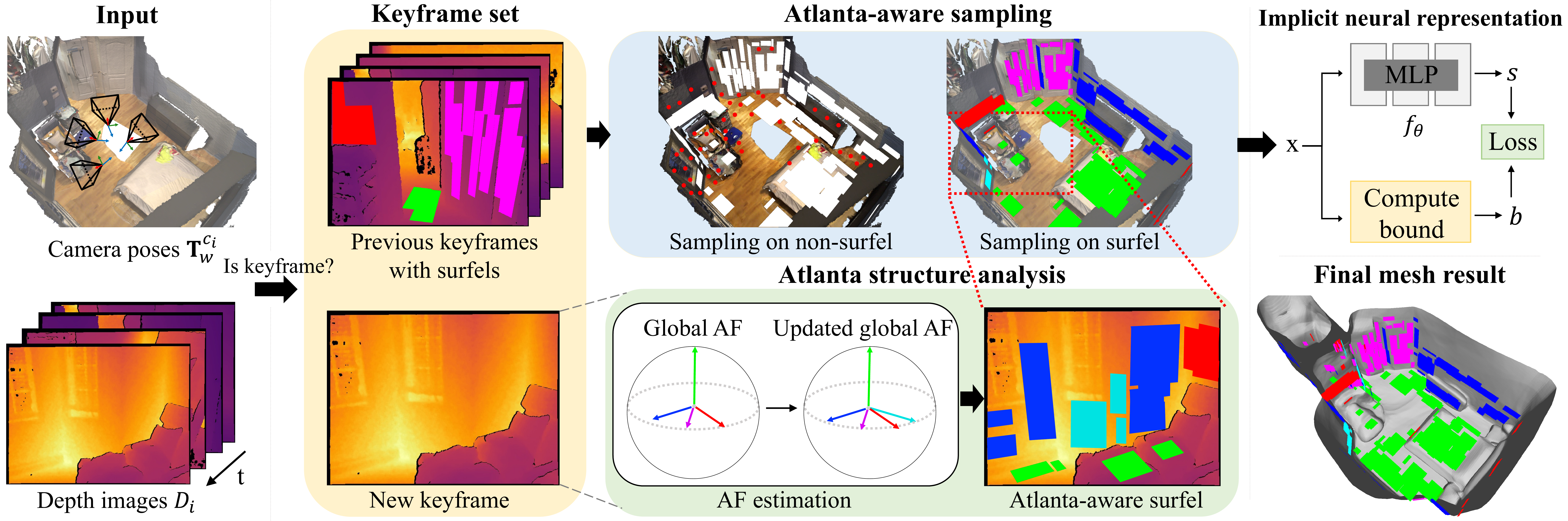}
    \captionof{figure}{
    \textbf{{Overview of {\AiSDF}}}.
    {Given a stream of posed depth images, AiSDF first selects the keyframe and adds it to the keyframe set for continual learning. 
    We update the global Atlanta frame (AF) by extracting the dominant directions from a new keyframe and then generate surfels that represent the planar regions supported by the updated global AF. 
    From a set of keyframes with Atlanta-aware surfels, we sample the 3D points considering the structure of the scene. 
    Finally, sampled point x is queried to MLP that outputs signed distance value s, and we optimize the network in a self-supervised manner by measuring the loss between s and bound b.
    Note that we intentionally present intermediate steps of continual learning to show the process of extracting the new Atlanta direction and surfels supported by updated global AF. 
    In Atlanta-aware sampling (blue box), we use the ground truth mesh to visualize the sampling effectively. 
    The final mesh result indicates the reconstructed mesh by AiSDF using all keyframes.}}
    \vspace{-4mm}
    \label{fig:overview}
\end{figure*}

Several research works~\cite{wang2022neuris,yu2022monosdf, guo2022neural} pay attention to enhancing the quality of neural scene reconstructions by combining geometric priors (\eg, depth and normal).\footnote{Note that this line of research works focuses on enhancing reconstruction quality by offline process regardless of computational efficiency.} 
Guo \etal~\cite{guo2022neural} propose a neural 3D scene reconstruction method with a strict structural assumption of indoor scenes.
This ManhattanSDF method seamlessly combines the planar regions following the MW assumption with the learning of implicit neural representations and improves the reconstruction quality, especially in textureless and homogeneous regions.
While the above methods show improved reconstruction quality, they cannot be utilized in the online process.
Therefore, we seamlessly integrate an online SDF reconstruction framework with a general structural assumption, the AW assumption.

\vspace{1mm} \noindent \textbf{Sampling strategy}. \
Unlike offline scene reconstruction approaches~\cite{Azinovic_2022_CVPR, murez2020atlas, guo2022neural} that fully utilize input data regardless of computational time and memory consumption, sampling strategy becomes essential for online scene reconstruction to ensure efficiency.
In the 2D image domain, iMAP~\cite{sucar2021imap} proposes loss-guided active sampling, which samples more points in the complex (high-frequency) regions based on loss calculated from the image grid. 
In the depth domain, iSDF~\cite{iSDF2022} randomly samples a small number of pixel coordinates (${\leq}200$) from each selected keyframe for efficiency.
Such sparse sampling strategies for scene reconstructions require an appropriate sampling rate based on the structural complexity of the scene. 
However, the accurate decision on the complexity of scenes in 2D or 2.5D domains is still a difficult challenge. 
To alleviate this issue, we exploit planar surfel regions supported by the AW assumption, which allows us to use Atlanta structure-aware sampling.

\section{Atlanta-aware iSDF Framework}  \label{sec:overview}

In this work, we propose a new structure-aware online SDF reconstruction framework in indoor scenes (see \Fref{fig:overview}).
Unlike the previous iSDF~\cite{iSDF2022} that focuses on reconstructing SDF itself, we exploit the structural regularity of indoor environments, especially the AW assumption~\cite{SchindlerDellaert04}.
We dub this structure-aware online SDF estimation {\AiSDF} in short.

{\AiSDF} takes as input a stream of posed depth images $\{D_i\}$ and camera poses $\{\mathbf{T}_{w}^{c_i}\}$, and aims to learn a neural network $f(\mathbf{x})$ based on the structural understanding of the AW, where $f(\mathbf{x})$ estimates SDF value $s$ at a 3D point $\mathbf{x} \in \mathbb{R}^3$.
Specifically, for a consecutively selected keyframe $\mathcal{K}_j{=}(D_j,\mathbf{T}_{w}^{c_j})$,
{\AiSDF} infers the underlying Atlanta structure (\ie,~Atlanta frame) within a continual framework. 
Based on this structural understanding, we then estimate planar regions that support the estimated AF in the form of surfel representation.
According to surfel regions, we perform Atlanta-aware sampling to force the structural regularity and focus on complex regions in SDF reconstruction adaptively.
Thus, the proposed {\AiSDF} framework improves the SDF reconstruction quality at the structure level as well as generates an explicit 3D planar map composed of Atlanta-aware surfels.

\subsection{Structural assumption: Atlanta frame}

The AW assumption~\cite{SchindlerDellaert04} can approximate a given indoor scene into a set of orthogonal and parallel planes, where planar walls are orthogonal to floors but do not have to be orthogonal to each other. 
We can formally define a set of dominant directions satisfying the AW assumption, which consists of a vertical dominant direction $\mathbf{v}_v$ and a set of $M$ horizontal dominant directions $\mathbf{v}_{h_m}$, where $\mathbf{v}_v \perp \mathbf{v}_{h_m}$. 
We call this direction set $\mathcal{V}=\{\mathbf{v}_v{,}\mathbf{v}_{h_1}{,}\mathbf{v}_{h_2}{,}{\cdots}{,}\mathbf{v}_{h_{M}}\}$ the \emph{Atlanta frame (AF)} or \emph{Atlanta directions}. 
In this work, we assume that a given indoor scene follows the AW assumption and use the AF parametrization~\cite{joo2019globally}\footnote{
    AF parametrization~\cite{joo2019globally} uses the rotation matrix~$\mathbf{R} = [\mathbf{r}_1,\mathbf{r}_2,\mathbf{r}_3] \in SO(3)$ to represent the vertical direction and the first horizontal direction (\ie, $\mathbf{v}_v = \mathbf{r}_1$ and $\mathbf{v}_{h_1}= \mathbf{r}_2$, where $\mathbf{v}_{h_1}$ acts as a reference location).
    Then, each $\mathbf{v}_{h_m}$ can be defined as a 1D angle parameter~$\alpha_m$ by rotating $\mathbf{v}_{h_1}$ by $\alpha_m$ around $\mathbf{v}_v$.
} that represents Atlanta directions using the rotation matrix $\mathbf{R}$ and a set of 1D angles $\{\alpha_m\}$.
We denote the estimated AF for the given $j$-th keyframe at the camera coordinate as local AF $\mathcal{V}_L^j$ and the unified AF for the observed keyframes at the world coordinate as global AF $\mathcal{V}_G$.

\begin{figure}[t]
    \centering
    \includegraphics[width=0.99\linewidth]{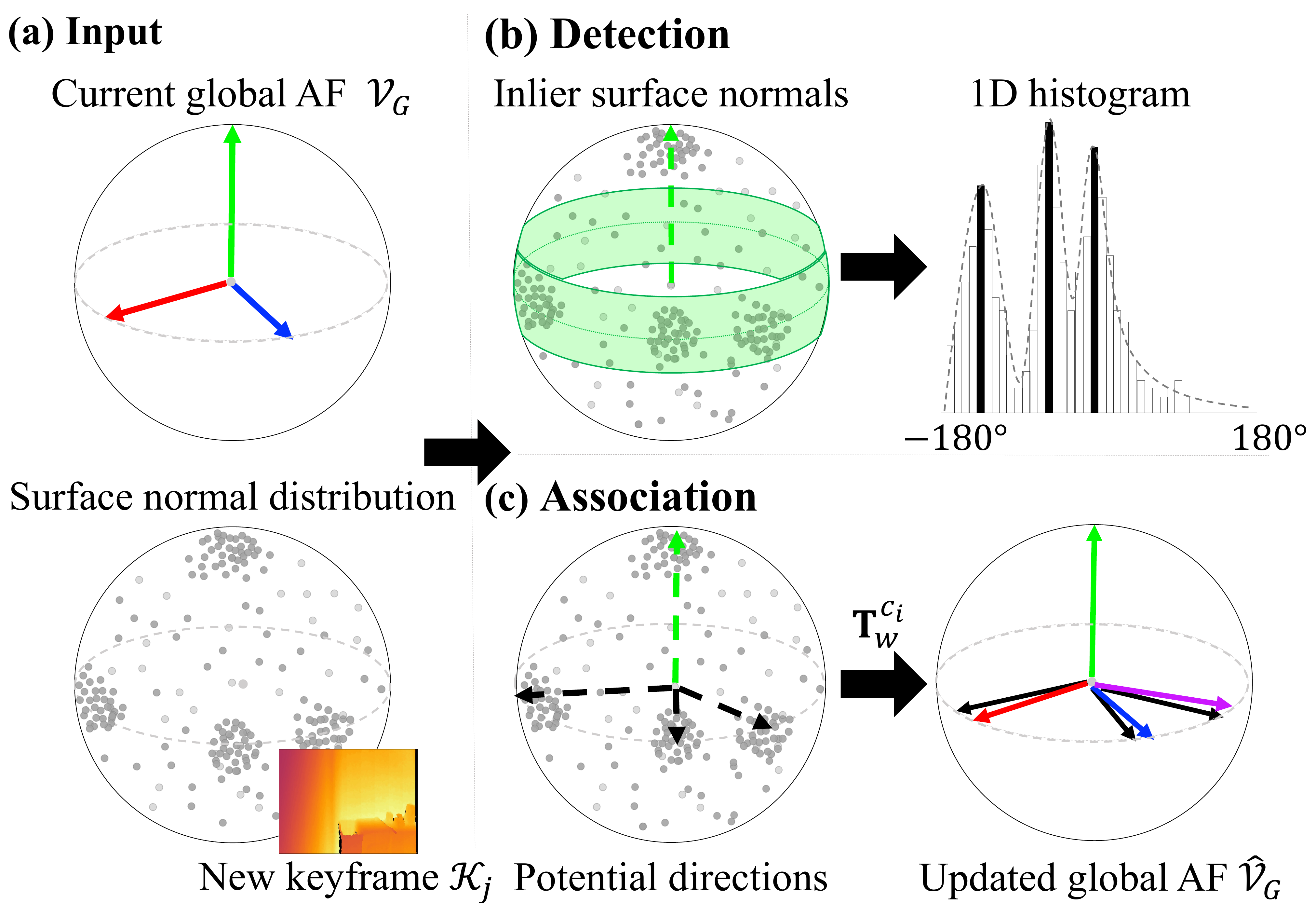}
    \caption{
        \textbf{Illustration of underlying AF estimation}. 
        (a) Given global AF $\mathcal{V}_G$ (solid arrows) and surface normal distribution of new keyframe $\mathcal{K}_j$, the Atlanta structure analysis proceeds in two steps. (b) First, we estimate potential dominant horizontal directions (black bars) from a 1D histogram of inlier surface normals. (c) The new dominant direction (purple arrows) is extracted by associating potential directions with the global AF in the world coordinate. 
    }
    \vspace{-4mm}
    \label{fig:AbyD}
\end{figure}

\subsection{Estimation of underlying Atlanta frame} \label{subsec:af_estimation}
For a consecutively selected keyframe, we estimate the underlying Atlanta structure (\ie, global AF) in a continual manner, as shown in \Fref{fig:AbyD}.
Unlike the previous work~\cite{joo2021linear} that detects the AF at each frame and then naively associates them, we fully exploit the property of the AW assumption that dominant horizontal directions are orthogonal to the vertical direction. 
That is, the potential horizontal direction does exist on the horizon defined by the vertical direction.
Based on this property, we reduce search space for the potential horizontal directions to 1D space in the range of [$-180^\circ$, $180^\circ$].
In this 1D domain, we detect dominant horizontal directions and then associate them with the current global AF.

Specifically, given the current global AF $\mathcal{V}_G$ and a new posed keyframe $\mathcal{K}_j$ with its surface normal distribution\footnote{
    Given a depth image and intrinsic parameter, we can directly compute the corresponding surface normal map using x-axis and y-axis gradient values and their cross-product.
 }, we first transform the vertical direction $\mathbf{v}_v {\in} \mathcal{V}_G$ into camera coordinate by ${\mathbf{T}_{w}^{c_j}}^{-1}$.
 We then set an inlier horizon region \wrt the vertical direction and identify the inlier surface normals within the inlier horizon region (see the green band region in \Fref{fig:AbyD}(b)). 
These inlier surface normals describe the distribution of potential dominant horizontal directions of the new keyframe. 
Thus, we construct a 1D histogram with 361 buckets, spaced at 1$^{\circ}$ intervals in the range [-180$^{\circ}$, 180$^{\circ}$] for the inlier surface normals, and detect the dominant directions from peak points of smoothed histogram via Gaussian filtering. 
By doing so, we can efficiently estimate potential dominant horizontal directions under a known vertical direction.
Finally, we transform the potential directions to the world coordinate by $\mathbf{T}_{w}^{c_j}$ and then associate them with the current global AF to determine the local AF $\mathcal{V}_{L}^j$ and update the global AF $\hat{\mathcal{V}}_G$ (see \Fref{fig:AbyD}(c)).

We initialize $\mathcal{V}_G$ as the dominant Manhattan directions (\eg,~\cite{joo2018robust}) and set the angle thresholds for inlier horizon region and association as $20^\circ$ to detect the dominant one and be robust against noise.

\subsection{Atlanta-aware surfel representation} \label{subsec:af_surfel}

\begin{figure}
	\centering
	\includegraphics[width=0.95\linewidth]{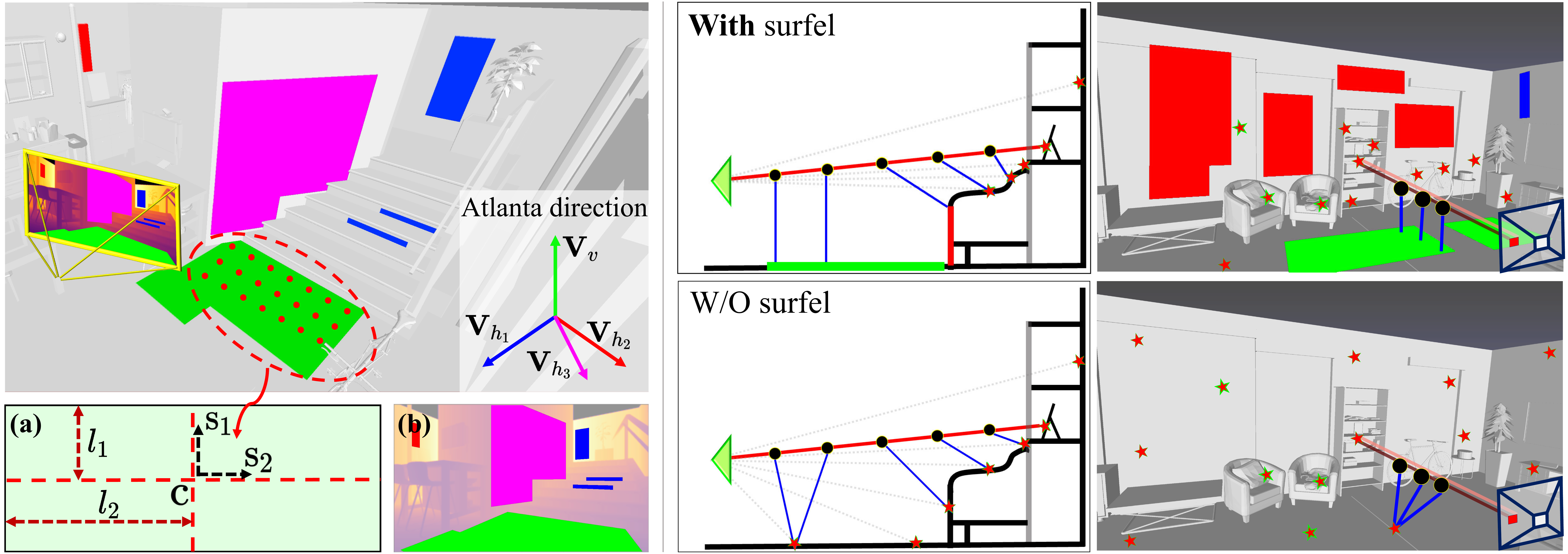}
	\caption{
	    \textbf{Illustration of the Atlanta-aware surfels and surfel-aware bound computation}. 
            \emph{Left}: (a) Atlanta-aware surfel representation. (b) 2D surfel mask $\mathbf{M}_\mathfrak{s}$ overlaid on the depth image.
            \emph{Right}: Selected points to compute bound values with surfels (top) and without surfels (bottom). 
            With Atlanta-aware surfels, {\AiSDF} can compute more tight bound values while covering a denser and wider area.
    }
    \label{fig:surfel_bound}
    \vspace{-3mm}
\end{figure}

Given a keyframe $\mathcal{K}_j$ with the estimated local AF $\mathcal{V}^j_L$, 
we extract dominant planar regions supporting the AF in the form of surface elements, \emph{Atlanta-aware rectangular surfels}, as shown in \Fref{fig:surfel_bound}(a).
We represent an Atlanta-aware rectangular surfel ${\mathfrak{s}}$ with its center point $\mathbf{c}$, surface normal $\mathbf{n}$, two axes $\mathbf{s}_1$ and $\mathbf{s}_2$, and the corresponding lengths $l_1$ and $l_2$.
Through Atlanta directions, we can constrain the surface normal of surfels and define axes of rectangular surfels using other Atlanta directions orthogonal to the surface normal.
This process allows us to explicitly extract a set of rectangular surfels without a learning-based module (\eg, wall segmentation module~\cite{guo2022neural}) and force the structural regularity inside the online SDF framework.

We first detect an Atlanta-aware plane using plane RANSAC as in \cite{joo2021linear}.
Let $\mathbf{v}$ be one of the given Atlanta directions and $\mathcal{X}_\mathbf{v}$ be a set of 3D points having similar normals with $\mathbf{v}$.
With randomly sampled 1-point in $\mathcal{X}_\mathbf{v}$ and $\mathbf{v}$, we can compute a potential Atlanta-aware plane and perform plane RANSAC using the Euclidean distance between the potential plane and $\mathcal{X}_\mathbf{v}$. 
We denote the detected Atlanta-aware planes by RANSAC as $\pi_\mathbf{v}$ and its inlier points as $\hat{\mathcal{X}}_\mathbf{v} {\subset} \mathcal{X}_\mathbf{v}$.
We then extract Atlanta-aware rectangular surfel ${\mathfrak{s}}$ on the detected plane, where its surface normal and two axes are directly computed from $\mathbf{v}$ and two orthogonal Atlanta directions (\eg, $\mathbf{n} = \mathbf{v}_v$, $\mathbf{s}_1=\mathbf{v}_{h_1}$, and $\mathbf{s}_2 {=} \mathbf{v}_{v} {\times} \mathbf{v}_{h_1}$). 
Specifically, we project $\hat{\mathcal{X}}_\mathbf{v}$ on $\pi_\mathbf{v}$, of which 2D space is defined by two axes $\mathbf{s}_1$ and $\mathbf{s}_2$.
On this 2D projection domain, we randomly sample 2-point and construct an axis-aligned rectangular surfel. 
We generate a set of candidate rectangular surfels by random sampling and select dominant surfels in terms of their area and occupancy by points, where the occupancy is measured by a ratio of associated points on the surfel against $\hat{\mathcal{X}}_\mathbf{v}$.
For each direction $\mathbf{v} \in \mathcal{V}^j_L$, we extract Atlanta-aware surfels $\{\mathfrak{s}\}$ in 3D space and 2D surfel mask $\mathbf{M}_\mathfrak{s}$ in the depth image domain. 

\subsection{Atlanta-aware sampling} \label{subsec:af_sampling}

Given a set of keyframes $\{\mathcal{K}_j\}$ with the estimated structure-aware surfels $\{\mathfrak{s}\}$, we perform structure-aware sampling that effectively considers planar surfel regions (\ie, 2D surfel mask on depth image domain).
According to the surfel mask $\mathbf{M}_\mathfrak{s}$, we separately sample points on the surfel and non-surfel regions ($N_\mathfrak{s}$ number of surfel points $\mathcal{X}_\mathfrak{s}{=}\{\mathbf{x}_\mathfrak{s} \}$ and $N_\mathfrak{n}$ number of non-surfel points $\mathcal{X}_\mathfrak{n}{=}\{\mathbf{x}_\mathfrak{n}\}$). 
By performing this structure-aware sampling, we intend to sample more non-surfel points (\ie, relatively dense sampling) on the complex non-surfel regions while fewer surfel points on homogeneous surfel regions, {as illustrated in the blue-colored box of \Fref{fig:overview}.}

Concretely, for the non-surfel regions, we randomly select a set of pixels and then sample a bunch of 3D points along the corresponding ray for each sample pixel as in~\cite{iSDF2022}. 
For the surfel regions, we uniformly sample 3D points on the 3D surfel region directly (see the left side of \Fref{fig:surfel_bound}), where the number of sampled points for each surfel is the same as the number of sampled points for a pixel along the ray for the non-surfel region. 

We then compute the bound and approximated gradient depending on where the points are sampled. 
In the case of bound computation, we use zero bound value for $\mathcal{X}_\mathfrak{s}$ and the closest distance to surface points and surfel points for $\mathcal{X}_\mathfrak{n}$, where we consider the sign by computing the difference of depth between the depth from a sensor and the point sample along the ray:
\begin{equation}
b(\mathbf{x}, \mathcal{P}) {=} 
    \begin{cases}
        \operatorname{sgn}(D[u, v]-d) \underset{\mathbf{p} \in \mathcal{P}} {\min}{| \mathbf{x}-\mathbf{p}|} & \mathbf{x}{\in} \mathcal{X}_\mathfrak{n} 
        \\ 0 & \mathbf{x} {\in} \mathcal{X}_\mathfrak{s} 
    \end{cases}
\label{eq:bound}
\end{equation}
where $\operatorname{sgn}(\cdot)$ is the sign function, $\mathcal{P}$ is a set of surface points and surfel points, and $D[u,v]$ and $d$ are depth values of pixel $[u,v]$ and query point, respectively.
The approximated gradient is computed with the closest surface point for $\mathbf{x} \in \mathcal{X}_\mathfrak{n}$, while it is replaced with the Atlanta direction for $\mathcal{X}_\mathfrak{s}$:
\begin{equation}
\hspace{-2.5mm}
g(\mathbf{x}, \mathcal{P}) {=} 
    \begin{cases}
        \operatorname{sgn}(D[u, v]{-}d) ( \mathbf{x}{-}\underset{\mathbf{p} \in \mathcal{P}}{\argmin}|\mathbf{x}{-}\mathbf{p}|) &\mathbf{x} {\in} \mathcal{X}_\mathfrak{n} 
        \\ \mathbf{v} &\mathbf{x} {\in} \mathcal{X}_\mathfrak{s}
    \end{cases}
\label{eq:grad}
\end{equation}
where $\mathbf{v}$ is an Atlanta direction. 
By directly sampling points from surfels, we can cover a denser and wider area without increasing the total number of samples.
This leads to a more accurate approximation of the SDF and gradient, as shown on the right side of \Fref{fig:surfel_bound}.
As a result, using these approximated values according to the type of sampled points, we can enforce \texttt{AiSDF} to learn more tight SDF reconstruction in a self-supervised manner.


\subsection{Loss for training {\AiSDF}} \label{subsec:loss_aisdf}
Similar to \cite{iSDF2022}, the loss function for training {\AiSDF} basically consists of three terms: SDF, gradient, and Eikonal loss terms to satisfy the geometric properties of SDF.
We adaptively utilize each loss term according to the identity of sampled 3D points (\ie, surfel points  $\mathcal{X}_\mathfrak{s} $ or non-surfel points  $\mathcal{X}_\mathfrak{n}$). 
In the case of non-surfel points as query points, we optimize the network using three loss functions used in iSDF (see details in~\cite{iSDF2022}) with approximated bound and gradient computed by Eqs.~(\ref{eq:bound}) and (\ref{eq:grad}).
In this work, we introduce a surfel-guided loss function for surfel points, which allows us to enforce structural regularity.

\vspace{1mm} \noindent \textbf{Surfel loss}. \
Given sampled surfel points $\mathbf{x}_\mathfrak{s} {\in} \mathcal{X}_\mathfrak{s}$, we strongly constrain planar regions. 
To represent its geometry, the predicted SDF value is forced to have zero value:
\begin{equation}
\mathcal{L}_{\text {sdf}}^{\mathfrak{s}}(f(\mathbf{x}_\mathfrak{s} ; \theta))=|(f(\mathbf{x}_\mathfrak{s} ; \theta)|.
\end{equation}
Also, since the normal of $\mathbf{x}_\mathfrak{s}$ is aligned with Atlanta direction $\mathbf{v}$, we can apply the gradient loss to align the two vectors:
\begin{equation}
    \mathcal{L}_{\text {grad}}^{\mathfrak{s}}(\nabla_{\mathbf{x}_\mathfrak{s}} f(\mathbf{x}_\mathfrak{s} ; \theta), \mathbf{v}) = 1-\frac{\nabla_{\mathbf{x}_\mathfrak{s}} f(\mathbf{x}_\mathfrak{s} ; \theta) \cdot \mathbf{v}}{\left\|\nabla_{\mathbf{x}_\mathfrak{s}} f(\mathbf{x}_\mathfrak{s} ; \theta)\right\|\|\mathbf{v}\|}, 
\end{equation}
where $\nabla_{\mathbf{x}_\mathfrak{s}} f(\mathbf{x}_\mathfrak{s} ; \theta)$ denotes the gradient of the predicted SDF.
In particular, we can constrain the normal in a common direction regardless of keyframes due to using global AF, which is one of our contributions using structural regularity.

Following~\cite{gropp2020implicit}, the Eikonal loss enforces the SDF to have a unit norm gradient to produce a high-fidelity surface:
\begin{equation}
\mathcal{L}_{\text {eik}}^{\mathfrak{s}}(f(\mathbf{x}_\mathfrak{s} ; \theta)) = 
        \left|\left\|\nabla_{\mathbf{x}_\mathfrak{s}} f(\mathbf{x}_\mathfrak{s} ; \theta)\right\|-1\right|.
\end{equation}

\noindent Our entire surfel loss is as follows: 
\begin{equation}
\mathcal{L}_\text{surfel}=\lambda_{\text {sdf}}^\mathfrak{s} \mathcal{L}_{\text{sdf}}^{\mathfrak{s}} + \lambda_{\text {grad}}^\mathfrak{s} \mathcal{L}_{\text{grad}}^{\mathfrak{s}} + \lambda_{\text {eik}}^\mathfrak{s} \mathcal{L}_{\text{eik}}^\mathfrak{s},
\end{equation}where $\lambda_{\text {sdf}}^\mathfrak{s}$, $\lambda_{\text {grad}}^\mathfrak{s}$ and  $\lambda_{\text {eik}}^\mathfrak{s}$ are the weight factors of the SDF loss, gradient loss and eikonal loss for surfel points.
We set $\{\lambda_{\text {sdf}}^\mathfrak{s}$, $\lambda_{\text {grad}}^\mathfrak{s}$, $\lambda_{\text {eik}}^\mathfrak{s}$\} = \{1, 0.4, 0.2\}.

\vspace{1mm}    \noindent \textbf{Total loss}. \ 
Finally, the total loss of {\AiSDF} is as follows:
\begin{equation}
l(\theta) =\mathcal{L}_{\text{iSDF}} + \mathcal{L}_{\text{surfel}},
\end{equation}
where $\mathcal{L}_\text{iSDF}$ indicates the loss term composed of SDF, gradient, and Eikonal loss for non-surfel sampled 3D points.
We set the weight factors of each loss term of $\mathcal{L}_\text{iSDF}$ to be the same as iSDF.

\section{Evaluation}

In this section, we demonstrate the proposed {\AiSDF} framework on synthetic and real-world datasets.
In \Sref{subsec:exp_setting}, we provide the details of the experiment setting. 
In \Sref{subsec:qual} and \Sref{subsec:quan}, we show the qualitative and quantitative results to give a better understanding of reconstruction in various scenarios. 
In addition, we analyze the proposed {\AiSDF} via an ablation study in \Sref{subsec:analysis}. 
It should be noted that additional experiments are available in the supplementary video.

\subsection{Experiment setting} \label{subsec:exp_setting}
\noindent \textbf{Implementation details}. \ 
Following \cite{iSDF2022}, we model {\AiSDF} as a single MLP composed of four hidden layers. 
Specifically, both sampled surfel points $\mathcal{X}_\mathfrak{s}$ and non-surfel points $\mathcal{X}_\mathfrak{n}$ are transformed by positional embedding to make the network learn high-frequency regions.
Then, embedded features are passed to four hidden layers that each composed of a linear layer and a softplus activation function.
The output of the network is the SDF value and produces a mesh from this value by running a marching cube algorithm.
{\AiSDF} is implemented by PyTorch~\cite{paszke2019pytorch} and trained on a single RTX 3090 GPU. 
To optimize our {\AiSDF}, we employ an AdamW optimizer~\cite{loshchilov2017decoupled} with a learning rate $1.3\times10^{-3}$ and a weight decay $1.2\times10^{-2}$. 
In addition, we utilize the same keyframes and iteration process as iSDF~\cite{iSDF2022} for a fair comparison. 

\begin{figure*}[!]
    \vspace{-3mm}
    \centering
    \captionsetup{type=figure}
    \includegraphics[width=0.85\linewidth]{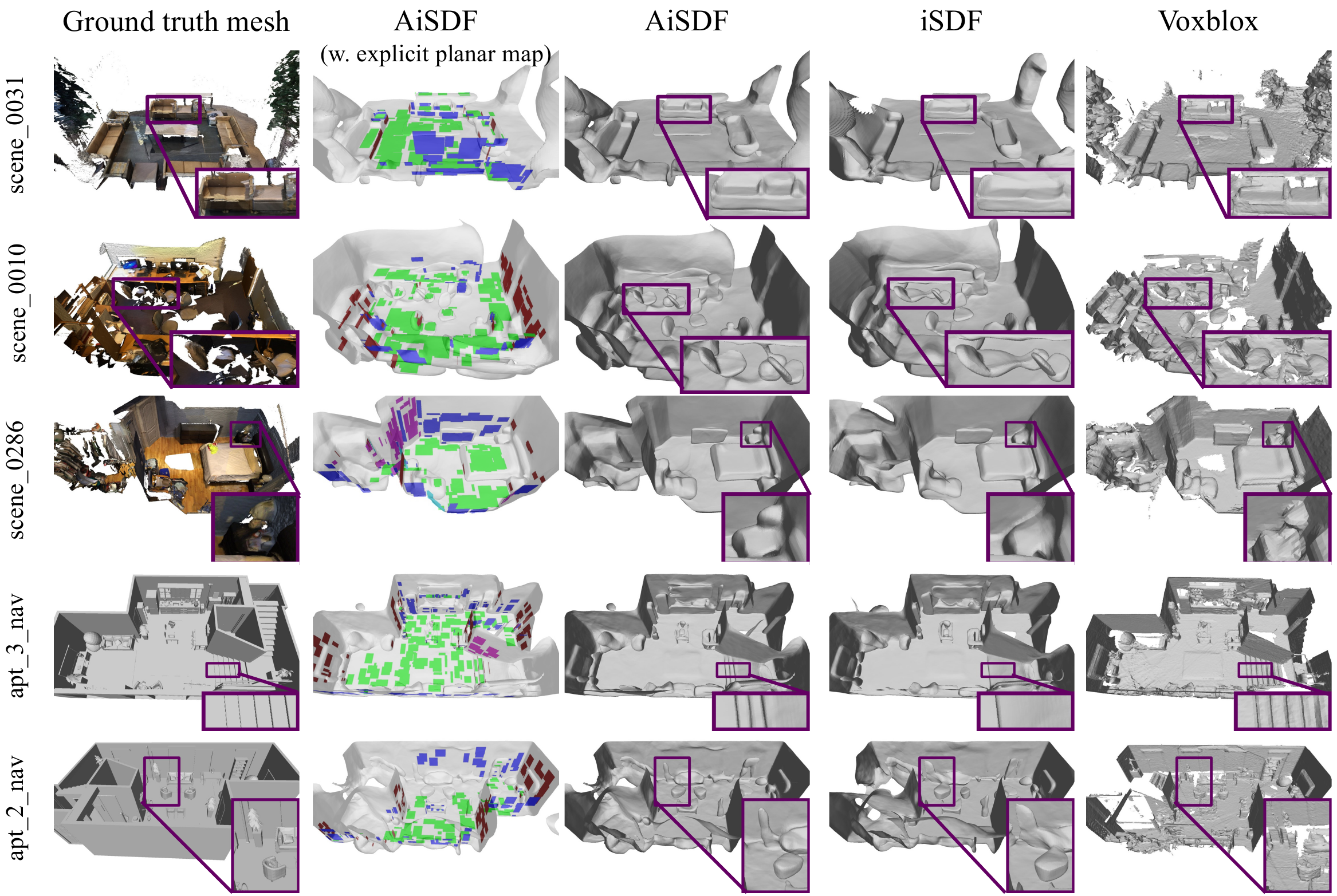}
    \captionof{figure}{
    \textbf{{Qualitative evaluation on the ScanNet and ReplicaCAD datasets}}. 
     The purple boxes are close-up views to show the details of each scene.
     The second column presents the half-transparent explicit planar map composed of surfels overlaid on the mesh.
     Here, the green color denotes the surfels supported by the vertical Atlanta direction, and the other colors represent the surfels by the other horizontal Atlanta directions.
     Note that the size of the surfels may look smaller in the ReplicaCAD due to the different scales of scenes. 
    }
    \label{fig:qualitative}
    \vspace{-3mm}
\end{figure*}

\vspace{1mm}\noindent \textbf{Dataset}. \
We validate the proposed {\AiSDF} on two datasets.
1) The ScanNet dataset~\cite{dai2017scannet} captured by an RGB-D camera contains $1,513$ scans of real-world indoor scenes with camera parameters, semantic labels, and surface reconstructions. 
Among these scans, we use six sequences satisfying the Manhattan structure and three sequences following the Atlanta structure.
2) The ReplicaCAD dataset~\cite{szot2021habitat} is a photo-realistic 3D indoor scene reconstruction dataset, which is a recreated version of the `FRL apartment' from the Replica dataset~\cite{replica19arxiv}. 
This dataset provides six scenes with variations of placement of large furniture, as well as small objects, for object rearrangement tasks. 
In our work, we use the same sequences used by iSDF~\cite{iSDF2022} for a fair comparison. 
Specifically, we utilize two rooms, where each room structure has three sequences according to different trajectories with their own purpose, such as navigation, object reconstruction, and manipulation (six sequences in total).

\vspace{1mm}\noindent \textbf{Comparison method}.
We compare {\texttt{AiSDF}} with two methods.
1) Voxblox~\cite{oleynikova2017voxblox} is an algorithm for incrementally propagating wavefronts from updated TSDF voxels to create a voxel grid with the Euclidean signed distance.
To conserve hardware resources, Voxblox uses voxels of large size and raycast grouping to accelerate integration. 
These techniques make it possible to create a map in real-time.
We set the voxel size to $5.5cm$.
2) iSDF~\cite{iSDF2022}, which is our baseline, is a real-time SDF reconstruction method from a stream of posed depth images of room-scale environments.

\vspace{1mm}\noindent \textbf{Metric}.\ 
For quantitative evaluation, we use the same three metrics as in iSDF~\cite{iSDF2022}: SDF error, collision cost error, and gradient cosine distance. 

\begin{figure}[!]
\centering
    \includegraphics[width=1\linewidth]{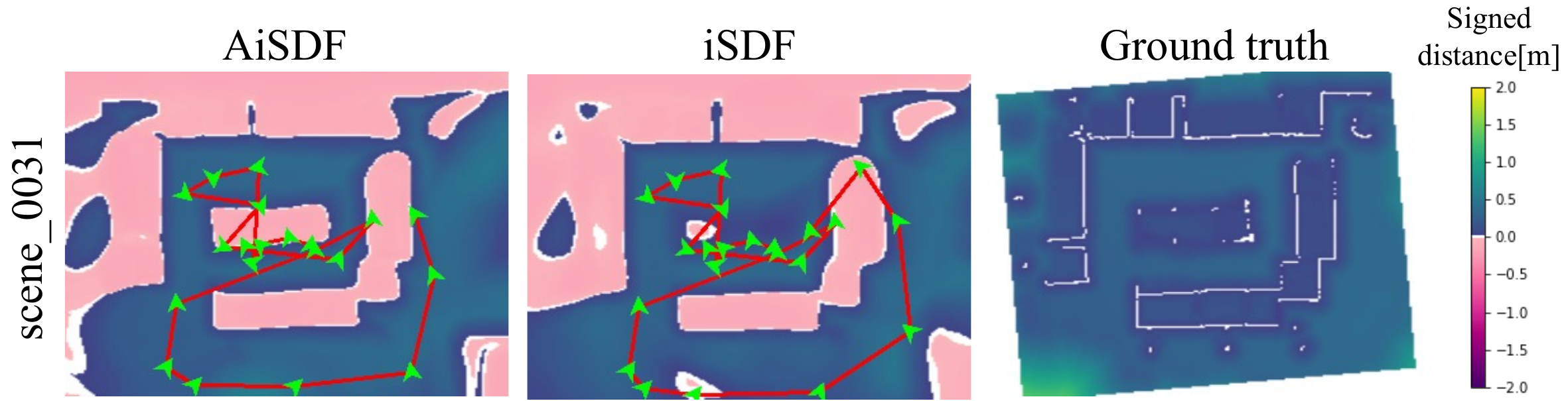}
    \caption{
        \textbf{Visualization of SDF at a constant height}. The SDF is extracted at the end of the sequence.
    }
    \vspace{-3mm}
    \label{fig:sdf_constant}
\end{figure}

\begin{table*}[t]
  \caption{\textbf{Quantitative evaluation}. 
  We highlight the best and second-best by \textbf{bold} and \underline{underline}, respectively. 
  }
  \vspace{-1mm}
  \label{[quantitative_results]}
  \centering
  \resizebox{0.95\linewidth}{!}{
    \begin{tabular}{cccccccccccccc}
    \toprule
    \multicolumn{2}{c}{} & \multicolumn{3}{c}{Voxblox~\cite{oleynikova2017voxblox}} & \multicolumn{3}{c}{iSDF~\cite{iSDF2022}} & \multicolumn{3}{c}{{\AiSDF} (w/o explicit map)} & \multicolumn{3}{c}{{\AiSDF} (w. explicit map)}\\
    \cmidrule(rl){3-5} \cmidrule(rl){6-8} \cmidrule(rl){9-11} \cmidrule(rl){12-14}
    \cmidrule(rl){3-5} \cmidrule(rl){6-8} \cmidrule(rl){9-11} \cmidrule(rl){12-14}
    \multicolumn{1}{c}{Dataset} & \multicolumn{1}{c}{Scene} & \multicolumn{1}{c}{SDF} & \multicolumn{1}{c}{Collision} & \multicolumn{1}{c}{Gradient} & \multicolumn{1}{c}{SDF} & \multicolumn{1}{c}{Collision} & \multicolumn{1}{c}{Gradient} & \multicolumn{1}{c}{SDF} & \multicolumn{1}{c}{Collision } & \multicolumn{1}{c}{Gradient} & \multicolumn{1}{c}{SDF $\downarrow$} & \multicolumn{1}{c}{Collision $\downarrow$} & \multicolumn{1}{c}{Gradient $\downarrow$}\\\hline
    \parbox[t]{3mm}{\multirow{6}{*}{\rotatebox[origin=c]{90}{ScanNet}}}
    & $scene\_0004$ & $6.352$ & $0.048$ & $0.215$ & $5.355$ & $0.037$ & $0.134$ & $\underline{4.317}$ & $\underline{0.029}$ & $\underline{0.125}$ & $\mathbf{4.258}$ & $\mathbf{0.028}$ & $\mathbf{0.125}$ \\
    & $scene\_0005$ & $6.416$ & $0.050$ & $0.120$ & $3.580$ & $0.029$ & $0.131$ & $\underline{3.470}$ & $\underline{0.028}$ & $\underline{0.127}$ & $\mathbf{3.402}$ & $\mathbf{0.028}$ & $\mathbf{0.124}$ \\
    & $scene\_0009$ & $5.608$ & $0.041$ & $0.112$ & $4.515$ & $0.034$ & $0.161$ & $\underline{3.726}$ & $\underline{0.028}$ & $\underline{0.149}$ & $\mathbf{3.593}$ & $\mathbf{0.027}$ & $\mathbf{0.145}$ \\
    & $scene\_0010$ & $4.850$ & $0.055$ & $0.347$ & $3.822$ & $0.032$ & $0.154$ & $\underline{3.494}$ & $\underline{0.029}$ & $\underline{0.149}$ & $\mathbf{3.285}$ & $\mathbf{0.027}$ & $\mathbf{0.147}$ \\
    & $scene\_0030$ & $5.500$ & $0.042$ & $0.942$ & $3.188$ & $0.026$ & $0.099$ & $\underline{2.910}$ & $\underline{0.023}$ & $\underline{0.095}$ & $\mathbf{2.734}$ & $\mathbf{0.022}$ & $\mathbf{0.092}$ \\
    & $scene\_0031$ & $11.512$ & $0.104$ & $0.317$ & $4.891$ & $0.036$ & $\mathbf{0.126}$ & $\underline{4.721}$ & $\underline{0.034}$ & $0.133$ & $\mathbf{4.646}$ & $\mathbf{0.034}$ & $\underline{0.132}$ \\\hline
    \parbox[t]{3mm}{\multirow{6}{*}{\rotatebox[origin=c]{90}{ReplicaCAD}}}
    & $apt\_2\_nav$ & $4.672$ & $0.038$ & $0.158$ & $4.349$ & $0.036$ & $0.156$ & $\underline{3.585}$ & $\underline{0.029}$ & $\underline{0.150}$ & $\mathbf{3.317}$ & $\mathbf{0.027}$ & $\mathbf{0.144}$ \\
    & $apt\_2\_mnp$ & $9.744$ & $0.073$ & $0.396$ & $7.707$ & $0.051$ & $0.339$ & $\underline{7.464}$ & $\underline{0.050}$ & $\underline{0.327}$ & $\mathbf{7.123}$ & $\mathbf{0.047}$ & $\mathbf{0.298}$ \\
    & $apt\_2\_obj$ & $8.178$ & $0.071$ & $0.237$ & $4.604$ & $0.034$ & $\underline{0.144}$ & $\underline{4.294}$ & $\underline{0.031}$ & $0.146$ & $\mathbf{4.119}$ & $\mathbf{0.030}$ & $\mathbf{0.143}$ \\
    & $apt\_3\_nav$ & $4.626$ & $0.036$ & $0.128$ & $3.713$ & $0.029$ & $0.119$ & $\underline{3.288}$ & $\underline{0.026}$ & $\underline{0.116}$ & $\mathbf{2.812}$ & $\mathbf{0.022}$ & $\mathbf{0.107}$ \\
    & $apt\_3\_mnp$ & $8.040$ & $0.076$ & $0.232$ & $6.336$ & $0.044$ & $\underline{0.192}$ & $\underline{5.979}$ & $\underline{0.042}$ & $0.194$ & $\mathbf{5.818}$ & $\mathbf{0.040}$ & $\mathbf{0.187}$ \\
    & $apt\_3\_obj$ & $5.802$ & $0.073$ & $0.282$ & $\underline{4.368}$ & $0.032$ & $\mathbf{0.133}$ & $4.406$ & $\underline{0.032}$ & $0.138$ & $\mathbf{4.294}$ & $\mathbf{0.031}$ & $\underline{0.134}$ \\
    \bottomrule
    \end{tabular}%
    }
\end{table*}

\subsection{Qualitative evaluation} \label{subsec:qual}

\noindent \textbf{{Mesh} reconstruction quality}. \ 
Figure~\ref{fig:qualitative} shows the mesh reconstruction results of {\AiSDF} and comparison methods.
Overall, Voxblox captures the details of even small objects, but it produces uneven surfaces. 
On the other hand, iSDF generates even surfaces, whereas has difficulty capturing the details of small objects.
Unlike comparison methods, the proposed {\AiSDF} reconstructs flat and complete surfaces while maintaining a certain level of detail thanks to Atlanta-aware sampling.
For example, {\AiSDF} reconstructs the complete shape of the sofa and the detail of the sofa armrest in $scene\_0031$ of ScanNet.
For more complex scenes, such as $scene\_0010$, we can observe that the shape and details of two nearby chairs are clearly distinguished by {\AiSDF}.
In the $apt\_3\_nav$ of ReplicaCAD, {\AiSDF} reconstructs the structure of stairs clearly compared to iSDF, which shows a collapsed structure.
Note that in the case of Voxblox, they show the clear structure of stairs since they use most of the input frames for reconstruction, unlike keyframe-based {\AiSDF} and iSDF (\ie,~the viewpoint covered by a few keyframes is limited). 
In addition, We can also produce the slice map of the reconstructed SDF, as shown in \Fref{fig:sdf_constant}.

\vspace{1mm}\noindent \textbf{Explicit planar map}. \
In addition to mesh reconstruction, {\AiSDF} can generate an explicit 3D planar map composed of Atlanta-aware surfels (see the second column in \Fref{fig:qualitative}).
In particular, explicit maps mainly contain planes of scene structure (\eg, walls and floors), while also involving relatively small planes (\eg, stairs and objects).
Specifically, even in $scene\_0010$ containing many objects, and $scene\_0286$ with Atlanta structure, \texttt{AiSDF} robustly extracts various planes supported by Atlanta directions. 
Furthermore, these explicit maps give additional geometric cues to reconstructed mesh.
For example, in $apt\_3\_nav$ of ReplicaCAD, {\AiSDF} reconstructs the structure of stairs at the mesh level by implicitly utilizing the extracted surfels in surfel loss.
We believe that explicit planar maps can also serve as an auxiliary resource for various downstream (\eg, footstep planners for humanoids~\cite{griffin2019footstep} and Roomplan scans~\cite{apple_roomplan}).
It should be noted that additional qualitative results are available in the supplementary video.

\subsection{Quantitative evaluation}  \label{subsec:quan}
\Tref{[quantitative_results]} shows the quantitative results. 
Following~\cite{iSDF2022}, for six scenes providing ground truth SDF, we measure three metrics for 200k sampled points on all rays computed by randomly sampled pixel coordinates from the frames.
In addition, we consider explicit planar maps together to measure the performance of {\AiSDF}.
To this end, we compute explicit SDF values by calculating the closest distance from each of the 200k sampled points to surfel maps.
Then, we compare it to the implicit SDF value and choose the smaller value to measure SDF and Collision metric with ground truth values.
For the sampled points using explicit SDF values, we replace implicit gradient vectors with explicit gradient vectors, obtained by computing the difference between the points and the corresponding surfel points to measure gradient metric.

Overall, {\AiSDF} outperforms comparison methods on the ScanNet and ReplicaCAD datasets. 
In particular, when we consider the estimated explicit planar map of {\AiSDF} together, it shows better performance.
Concretely, in the case of Voxblox, since it uses large voxel size for real-time, its performance is much worse than iSDF and {\AiSDF}.
Regarding iSDF, it is difficult to predict an accurate SDF because of the noisy depth of homogeneous regions.
In addition, iSDF may have difficulty assigning the proper number of samples to complex regions, whereas \texttt{AiSDF} samples more points on complex regions by utilizing the Atlanta-aware surfels.
On the other hand, {\AiSDF} obtains more accurate results by using Atlanta-aware surfels that are robust to noise and allow us adaptive sampling according to surfels or non-surfels.

\begin{table}[t]
    \caption{\textbf{Ablation study of Atlanta-aware sampling}.}
    \label{[ablation]}
    \vspace{-1mm}
    \centering
    \resizebox{0.95\columnwidth}{!}{
  \Large
    \begin{tabular}{cccccccc}
    \toprule
    \multicolumn{2}{c}{} & \multicolumn{3}{c}{$scene\_0010$} & \multicolumn{3}{c}{$apt\_2\_nav$}\\
    \cmidrule(rl){3-5} \cmidrule(rl){6-8}
    \multicolumn{1}{c}{Surfel mask} & \multicolumn{1}{c}{Surfel loss} & \multicolumn{1}{c}{SDF} & \multicolumn{1}{c}{Collision} & \multicolumn{1}{c}{Gradient} & \multicolumn{1}{c}{SDF} & \multicolumn{1}{c}{Collision} & \multicolumn{1}{c}{Gradient}\\\hline
    $\times$ & $\times$ & $3.822$ & $0.032$ & $0.154$ & $4.349$ & $0.036$ & $0.156$\\
    \checkmark & $\times$ & $3.667$ & $0.030$ & $0.154$ & $4.042$ & $0.033$ & $0.152$\\
    $\times$ & \checkmark & $3.760$ & $0.031$ & $0.151$ & $3.959$ & $0.032$ & $0.155$\\
    \checkmark & \checkmark & $\mathbf{3.494}$ & $\mathbf{0.029}$ & $\mathbf{0.149}$ & $\mathbf{3.585}$ & $\mathbf{0.029}$ & $\mathbf{0.150}$\\
    \bottomrule
    \end{tabular}%
    }
\end{table}

\subsection{Analysis} \label{subsec:analysis}
\noindent \textbf{Ablation study}. 
We conduct the ablation study to demonstrate the effectiveness of Atlanta-aware sampling: surfel mask and surfel loss (see \Tref{[ablation]}). 
Surfel mask $\mathbf{M}_\mathfrak{s}$ provides a criterion for dividing a scene into complex and planar regions.
Thus, training the model with surface masks allows us to sample more points on complex areas, resulting in improved performance.
In addition, we observe that regularizing the network with surfel loss leads to performance improvement.
Specifically, by utilizing surfel points $\mathcal{X}_\mathfrak{s}$, we can directly constrain the planar regions and obtain a more accurate approximated bound and gradient.
Consequently, {\AiSDF} using surface mask and surfel loss together shows the best performance.

\begin{table}    
    \caption{
        \textbf{Runtime of {\AiSDF}}.
        For the sequences of ScanNet used in the quantitative evaluation, we compute the average time for each module and then round up (unit: ms).
    }
    \vspace{-1mm}
    \label{[time_aisdf]}
    \centering
    \resizebox{0.95\columnwidth}{!}{
        \begin{tabular}{ccccccc}
        \hline
        AF estimation & Extract surfels & Sampling & Compute bound & forward & backward & \textbf{Total}  \\
        \hline
        $2$ & $54$ & $5$ & $7$ & $6$ & $14$ & $\mathbf{88}$ \\ \hline
        \end{tabular}%
    }
        \vspace{-3mm}
\end{table}

\vspace{1mm} \noindent \textbf{Runtime and memory}. \
\Tref{[time_aisdf]} shows  the running time for each module in \AiSDF.
Unfortunately, {\AiSDF} does not work in real-time, but it is sufficient to reconstruct the scene in an online process ($\le100ms$) when we consider this process that corresponds to the back-end part and takes only keyframes, not each frame.
In terms of memory, {\AiSDF} reconstructs the scene with 1MB of network parameters as in iSDF, while effectively representing the scene with only hundreds of KB of the explicit planar map.

\vspace{1mm} \noindent \textbf{Limitation}. \
{\AiSDF} proposes a novel way to combine the structural regularity of Atlanta structures with the implicit SDF reconstruction framework working online, which is our main contribution. 
Contrary to this novelty, {\AiSDF} has several limitations as a pioneer work.
To maintain reasonable runtime, {\AiSDF} independently extracts Atlanta-aware surfels for each keyframe.
Thus, we cannot provide a complete and unified explicit planar map of a given indoor scene. 
In addition, the current {\AiSDF} does not fully exploit the estimated Atlanta structures in the implicit neural representation; we currently enforce surfel loss on surfel points to learn SDF.
From this point of view, we believe that encoding Atlanta-aware surfels itself to SDF could be an interesting research direction.

\section{Conclusion} 
\label{sec:conclusion}
Under the AW assumption, we have proposed the novel {\AiSDF}, a structure-aware online SDF reconstruction framework of a given indoor scene. 
To fully exploit the inherent property of indoor scenes (\ie, structural regularity), we estimate the underlying Atlanta structure in the form of dominant Atlanta directions within a continual framework.  
This structural understanding allows us to extract Atlanta-aware surfels, which explicitly play a role as a 3D planar map.
Moreover, Atlanta-aware surfels provide a criterion to adaptively sample points and make \texttt{AiSDF} implicitly enforce surfel loss on surfel points.
We seamlessly integrate this structural understanding inside the online SDF reconstruction framework.    
As a result, experiments demonstrate that {\AiSDF} can reconstruct the details of the scene with overall structure while extracting the lightweight explicit planar map.

\bibliographystyle{IEEEtran}
\bibliography{egbib}

\end{document}